\documentclass[10pt]{article}

\usepackage{fullpage}
\usepackage{setspace}
\usepackage{parskip}
\usepackage{titlesec}
\usepackage[section]{placeins}
\usepackage{xcolor}
\usepackage{breakcites}
\usepackage{lineno}
\usepackage{hyphenat}
\usepackage{multirow}
\usepackage{color}
\newcommand{\myRed}[1]{\textcolor{black}{#1}}
\newcommand{\myRedRed}[1]{\textcolor{black}{#1}}

\PassOptionsToPackage{hyphens}{url}
\usepackage[colorlinks = true,
            linkcolor = blue,
            urlcolor  = blue,
            citecolor = blue,
            anchorcolor = blue]{hyperref}
\usepackage{etoolbox}
\makeatletter
\patchcmd\@combinedblfloats{\box\@outputbox}{\unvbox\@outputbox}{%
  \errmessage{\noexpand\@combinedblfloats could not be patched}%
}%
\makeatother

\usepackage{amsmath}

\renewenvironment{abstract}
  {{\bfseries\noindent{\abstractname}\par\nobreak}\footnotesize}
  {\bigskip}

\titlespacing{\section}{0pt}{*3}{*1}
\titlespacing{\subsection}{0pt}{*2}{*0.5}
\titlespacing{\subsubsection}{0pt}{*1.5}{0pt}

\usepackage{authblk}
\usepackage{hyperref}

\usepackage{graphicx}
\usepackage[space]{grffile}
\usepackage{latexsym}
\usepackage{textcomp}
\usepackage{longtable}
\usepackage{tabulary}
\usepackage{booktabs,array,multirow}
\usepackage{amsfonts,amsmath,amssymb}
\usepackage{subcaption}
\providecommand\citet{\cite}
\providecommand\citep{\cite}

\newcommand*\samethanks[1][\value{footnote}]{\footnotemark[#1]}
\newif\iflatexml\latexmlfalse

\AtBeginDocument{\DeclareGraphicsExtensions{.pdf,.PDF,.eps,.EPS,.png,.PNG,.tif,.TIF,.jpg,.JPG,.jpeg,.JPEG}}

\usepackage[utf8]{inputenc}
\usepackage[english]{babel}

\usepackage{float}
\usepackage{xcolor}

\usepackage[margin=1.5in]{geometry}
\DeclareMathOperator*{\argmin}{argmin}

\begin{document}

\title{Explainable Deep Learning in Healthcare: A Methodological Survey from an Attribution View [Advanced Review]}

\author[]{Di Jin\thanks{Equal contributions}}%
\author[]{Elena Sergeeva\samethanks}
\author[]{Wei-Hung Weng\samethanks}
\author[]{Geeticka Chauhan\samethanks}
\author[]{Peter Szolovits\thanks{Corresponding author: psz@mit.edu}}
\affil[]{Computer Science and Artificial Intelligence Laboratory\\ Massachusetts Institute of Technology, Cambridge, MA, 02139, USA}%

\vspace{-1em}

  \date{}

\begingroup
\let\center\flushleft
\let\endcenter\endflushleft
\maketitle
\endgroup

\selectlanguage{english}
\begin{abstract}
The increasing availability of large collections of electronic health record (EHR) data and unprecedented technical advances in deep learning (DL) have sparked a surge of research interest in developing DL based clinical decision support systems for diagnosis, prognosis, and treatment. Despite the recognition of the value of deep learning in healthcare, impediments to further adoption in real healthcare settings remain due to the black-box nature of DL. Therefore, there is an emerging need for interpretable DL, which allows end users to evaluate the model decision making to know whether to accept or reject predictions and recommendations before an action is taken. In this review, we focus on the interpretability of the DL models in healthcare. We start by introducing the methods for interpretability in depth \myRedRed{and comprehensively as a methodological reference} for future researchers or clinical practitioners in this field. Besides the methods' details, we also include a discussion of advantages and disadvantages of these methods and which scenarios each of them is suitable for, so that interested readers can know how to compare and choose among them for use. Moreover, we discuss how these methods, originally developed for solving general-domain problems, have been adapted and applied to healthcare problems and how they can help physicians better understand these data-driven technologies. Overall, we hope this survey can help researchers and practitioners in both artificial intelligence (AI) and clinical fields understand what methods we have for enhancing the interpretability of their DL models and choose the optimal one accordingly.
\end{abstract}%


This article is categorized under:

Keywords: Interpretable Deep Learning, Deep Learning in medicine


\sloppy

\section{Introduction}


In recent years, the wide adoption of electronic health record (EHR) systems by healthcare organizations and subsequent availability of large collections of EHR data have made the application of Artificial Intelligence (AI) techniques in healthcare more feasible. 
The EHR data contain rich, longitudinal, and patient-specific information including both structured data (e.g., patient demographics, diagnoses, procedures) as well as unstructured data, such as physician notes and medical images~\citep{mesko2017role}. Meanwhile, deep learning (DL), a family of machine learning (ML) models based on deep neural networks, has achieved remarkable progress in the last decade on various datasets for different modalities including images, natural language, and structured time series data~\citep{lecun2015deep}. The availability of large-scale data and unprecedented technical advances have come together to spark a surge of research interest in developing a variety of deep learning based clinical decision support systems for diagnosis, prognosis and treatment~\citep{murdoch2013inevitable}. 

Despite the recognition of the value of deep learning in healthcare, impediments to further adoption in real healthcare settings remain~\citep{Tonekaboni2019WhatCW}. One pivotal impediment relates to the \textit{black box} nature, or opacity, of deep learning algorithms, in which there is no easily discernible logic connecting the data about a case to the decisions of the model. Healthcare abounds with possible ``high stakes” applications of deep learning algorithms: 
predicting a patient’s likelihood of readmission to the hospital~\citep{ashfaq2019readmission}, making the diagnosis of a patient's disease~\citep{esteva2017dermatologist}, suggesting the optimal drug prescription and therapy plan~\citep{rough2020predicting}, just to name a few. In these critical use cases that include clinical decision making, there is some hesitation in the deployment of such models because the cost of model mis-classification is potentially high~\citep{mozaffari2014systematic}. Moreover, it has been widely demonstrated that deep learning models are not robust and may easily encounter failures in the face of both artificial and natural noise~\citep{Szegedy2014Intriguing,finlayson2019adversarial,Jin2020IsBR}.

Artificial intelligence (AI) systems are, on the whole, not expected to act autonomously in patient care, but to serve as decision support for human clinicians. To support the required communication between such systems and people, and to allow the person to assess the reliability of the system's advice, we seek to build systems that are interpretable. Interpretable
DL allows algorithm designers to interrogate, understand, debug, and even improve the systems to be deployed \myRed{by analyzing and interpreting the behavior of black-box DL systems.} From the end user perspective, interpretable DL allows end users to evaluate the model decision making to determine whether to accept or reject predictions and recommendations before an action is taken. 

In particular, in this review we focus on the interpretability of the DL models in health care. Such models are known for both their complexity and high performance on a variety of tasks, yet the decisions and recommendations of deep learning systems may be biased~\citep{gianfrancesco2018potential}. Interpretability can offer one effective approach to ensuring that such systems are free from bias and fair in scoring different ethnic and social groups~\citep{hajian2016algorithmic}. Many DL systems have already been deployed to make decisions and recommendations in non-healthcare settings for tens of millions of people around the world (e.g., Netflix, Google, Amazon) and we hope that DL applied in healthcare will also become widespread~\citep{esteva2019guide}. To this end, we need help from interpretability to better understand the resulting models to help prevent potential negative impacts. Lastly, there are some legal regulations such as the European Union (EU)’s General Data Protection Regulation (GDPR) that require organizations that use patient data for predictions and recommendations to provide on demand explanations \myRed{for an output of the algorithm, which is called a ``right to explanation''~\citep{tesfay2018read,edwards2018enslaving}.} The inability to provide such explanations on demand may result in large penalties for the organizations involved.

It should be noted that the notion of explanation of a decision in itself is not a very well defined concept: indeed the original EU GDPR Recital 71 does not provide a clear definition beyond stating a person's right to obtain it. \myRed{There have been active discussions in the community on this notion~\citep{lipton2018mythos}; for instance, \cite{muggleton2018ultra} proposed an operational definition of comprehensibility and interpretability based on the ultra-strong criteria for Machine Learning proposed by \cite{michie1988machine} and Inductive Logic Programming~\citep{kovalerchuk2021survey}. However, no single uniform definition has been reached:} for any complex model with no superficial components, any simple explanation is inherently unfaithful to the underlying model. The decision on what definition of explanation to use necessarily affects the properties of the methods used to produce them: for example focusing on producing a per example explanation vs.\ the structure of the network analysis favors local (why this particular example resulted in a given prediction) explanation over global ones (what kinds of knowledge are encoded in the model and how they affect predictions).  \myRed{In this work, we first cover the most common type of interpretation method, in which} an explanation is an assignment of a score to each input element that reflects its importance to a model's conclusions.  We also briefly discuss example based explanation methods. Other approaches to interpretability include a more recent focus on feature interactions for neural networks \citep{sundararajan2020shapley, tsang2018detecting,  tsang2020does} and whole network behavior analysis \citep{carter2019activation}.

It is conventionally thought that there is a trade-off between model interpretability and performance (e.g., F1, accuracy). For example, more interpretable models such as regression models and decision trees often perform less well on many prediction tasks compared to less interpretable models such as deep learning models. With this constraint, researchers have to balance the desire for the most highly performing model against adequate interpretability. Fortunately in the last few years, researchers have proposed many new methods that can maintain the model performance while producing good explanations, such as LIME~\citep{ribeiro2016should}, RETAIN~\citep{choi2016retain}, and SHAP~\citep{lundberg2017unified}, described below. And many of them have been adapted and applied to healthcare problems with good interpretability achieved. This survey aims to provide a comprehensive and in-depth summary and discussion over such methods.

Previous surveys on explainable ML for healthcare~\citep{ahmad2018interpretable,holzinger2019causability,wiens2019no,Tonekaboni2019WhatCW,vellido2019importance,payrovnaziri2020explainable} mainly discuss the definition, concept, importance, application, evaluation, and high-level overview of methods for interpretability. In contrast, we will focus on introducing the methods for interpretability in depth so as to provide methodological guidance for future researchers or clinical practitioners in this field. Besides the methods' details, we will also include a discussion of advantages and disadvantages of these methods and which scenarios each of them is suitable for, so that interested readers can know how to compare and choose among them for use. Moreover, we will discuss how these methods originally developed for solving general-domain problems have been adapted and applied to healthcare problems and how they can help physicians better understand these data-driven technologies. Overall, we hope this survey can help researchers and practitioners in both AI and clinical fields understand what methods we have for enhancing the interpretability of their DL models and choose the optimal one accordingly based on a deep and thorough understanding. For readers' convenience, we have provided a map between all abbreviations to be used and their corresponding full names in Table \ref{tab:glossary}.

\paragraph{Paper Selection:}
We first conducted a systematic search of papers using MEDLINE, IEEE Xplore, Association for Computing Machinery (ACM), and ACL Anthology databases, several prestigious clinical journals' websites such as Nature, JAMA, JAMIA, BMC, \myRed{Elsevier, Springer,} Plos One, etc., as well as the top AI conferences such as NeurIPS, ICML, ICLR, AAAI, KDD, etc.. The keywords for our searches are: (explainable OR explainability OR interpretable OR interpretability OR understandable OR understandability OR comprehensible OR comprehensibility) AND (machine learning OR artificial intelligence OR deep learning OR AI OR neural network). After initial searching, we conducted manual filtering by reading titles and abstracts and only retained three types of works for subsequent careful reading: interpretability methods developed for general domain problems, 
interpretability methods specifically developed for healthcare problems, 
and healthcare applications that involve interpretability. 
We only covered the methods that can interpret DL models. 
\myRed{The literature of explanation methods for DL grows rapidly, so any review of this type is captive to its date of completion. Searching the above-mentioned sources with the keywords we used for recent articles should help to bring an appreciation of the field up to date.}

\section{Interpretability Methods}

In this section, we will introduce various kinds of interpretability methods, which aim to assign an attribution value, sometimes also called "relevance" or "contribution", to each input feature of a network. Such interpretability methods can thus be called attribution methods. More formally, consider a deep neural network (DNN) that takes an input $x= [x_1,...,x_N]$ and produces an output $S(x) = [S_1(x),...,S_C(x)]$, where $C$ is the total number of output neurons. Given a specific target neuron $c$, the goal of an attribution method is to determine the contribution $R^c= [R^c_1,...,R^c_N]$ of each input feature $x_i$ to the output $S_c$. For a classification task, the target neuron of interest is usually the output neuron associated with the correct class for a given sample. 
The obtained attribution maps are usually displayed as heatmaps, where one color indicates features that contribute positively to the activation of the target output while another color indicates features that have a suppressing effect on it.

To organize our presentation, we classify all attribution methods into the following categories: back-propagation based, attention based, feature perturbation based, model distillation based, and game theory based. We also include example and generative based interpretation for DL methods for completeness. More technical details for each category will be elaborated below.

\subsection{Back-propagation}
\label{sec:back-propagation-methods}

\begin{figure*}[t!]
    \centering
    \begin{subfigure}[t]{0.5\textwidth}
        \centering
        \includegraphics[height=1.2in]{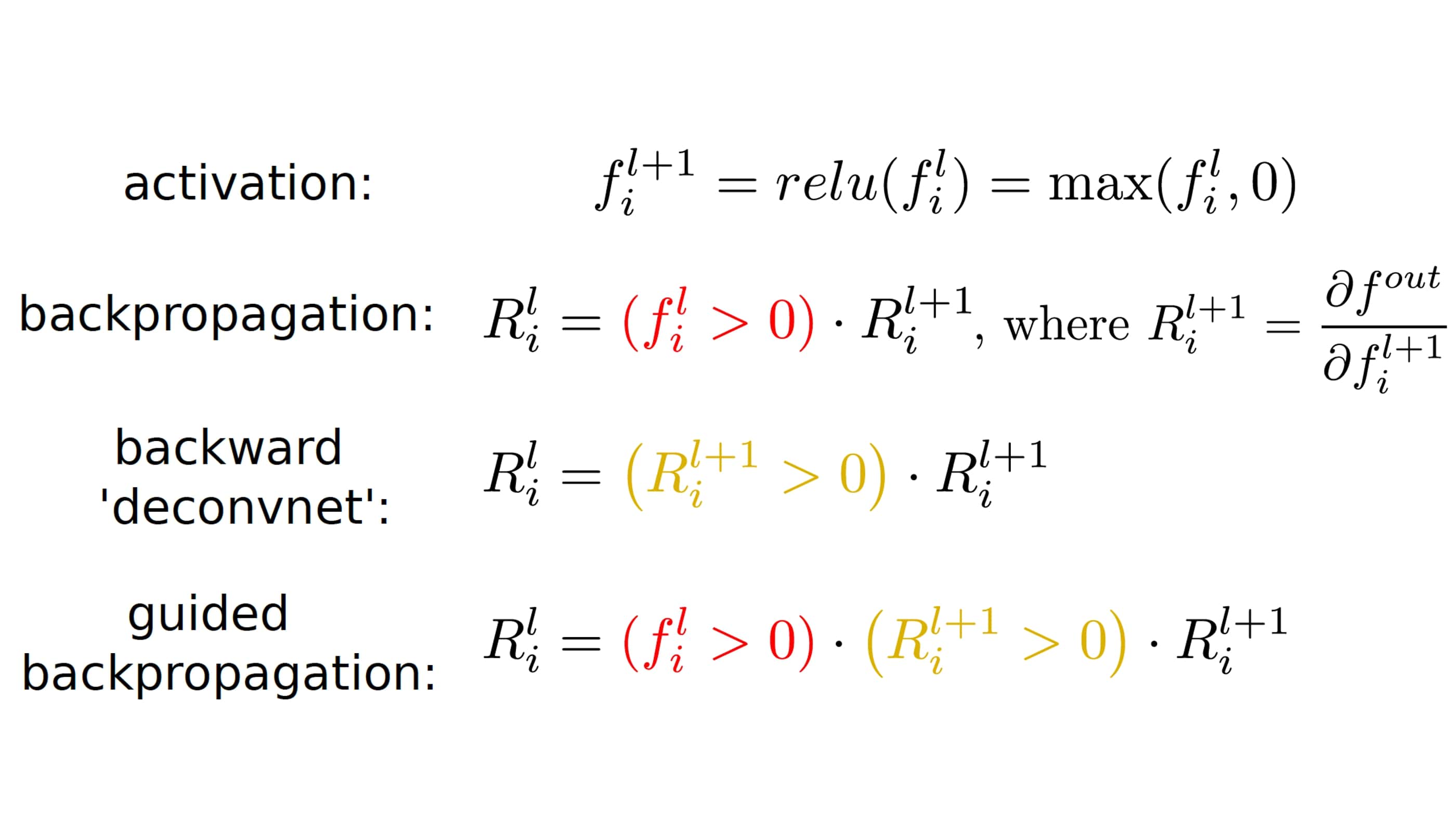}
        \caption{Comparison among normal gradient back-propagation, Deconvolution, and Guided Back-propagation in terms of how they handle back-propagation through the rectified linear (ReLU) non-linearity.}
        \label{figs:guided-back-propagation}
    \end{subfigure}%
    ~ 
    \begin{subfigure}[t]{0.5\textwidth}
        \centering
        \includegraphics[height=1.2in]{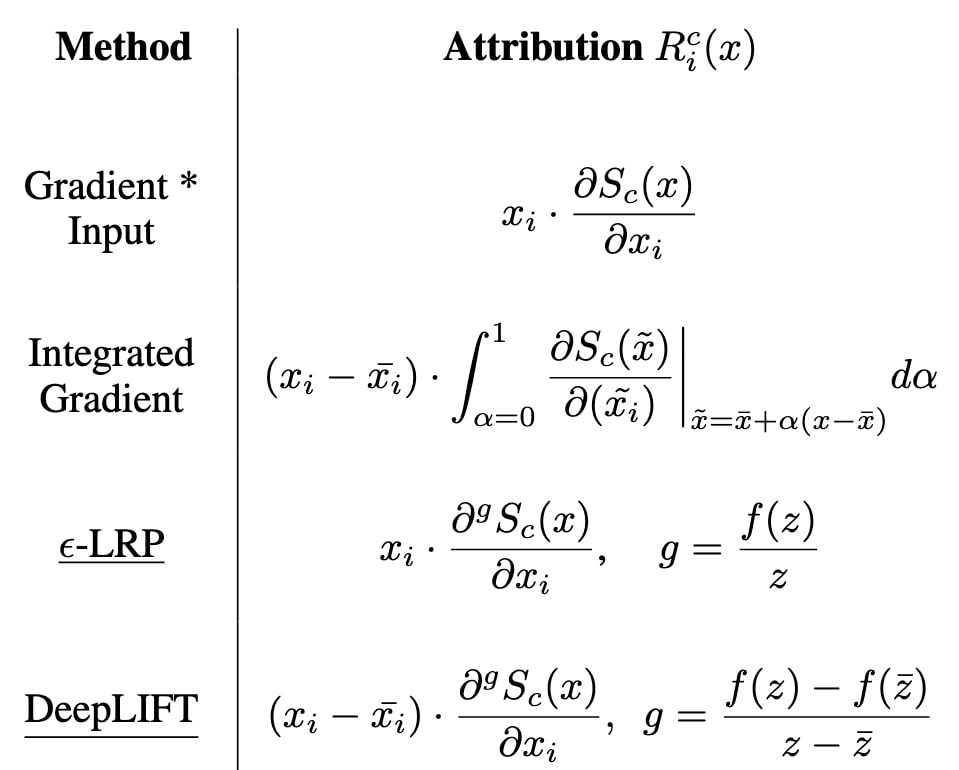}
        \caption{Mathematical formulation of four back-propagation based attribution methods. The original equations of $\epsilon$-LRP and DeepLIFT are transformed so that they can be calcuated based on gradients.}
        \label{figs:various-back-propagation}
    \end{subfigure}
    \caption{Mathematical formulation of different back-propagation based interpretability methods.}
\end{figure*}

The most popularly used interpretability method is based on back-propagation of either gradients~\citep{simonyan2013deep} or activation values~\citep{bach2015pixel}. This line of methods starts from the {Saliency Map}~\citep{simonyan2013deep}, which follows the normal gradient back-propagation process and constructs attributions by taking the absolute value of the partial derivative of the target output $S_c$ with respect to the input features $x_i$, i.e., $|\frac{\partial S_c(x)}{\partial x_i}|$. Intuitively, the absolute value of the gradient indicates those input features that can be perturbed the least in order for the target output to change the most. However, the absolute value prevents the detection of positive and negative evidence that might be present in the input. To make the reconstructed heatmaps significantly more accurate for convolutional neural network (CNN) models, {Deconvolution}~\citep{zeiler2014visualizing} and {Guided Back-propagation}~\citep{springenberg2014striving} were proposed and these two methods and the Saliency Map method differ mainly in the way they handle back-propagation through the rectified linear (ReLU) non-linearity. As illustrated in Figure \ref{figs:guided-back-propagation}, for normal gradient back-propagation in the Saliency Map method, when the activation values in the lower layer are negative, the corresponding back-propagated gradients are masked out. In contrast, the Deconvolution method masks out the gradients when they themselves are negative, while the Guided Back-propagation approach combines these two methods: those gradients are masked out for which at least one of these two values is negative. 

{Gradient * Input}~\citep{shrikumar2016not} was proposed as a technique to improve the sharpness of the attribution maps. The attribution is computed taking the (signed) partial derivatives of the output with respect to the input and multiplying them with the input itself. 
{Integrated Gradients}~\citep{sundararajan2017axiomatic} is similar to Gradient * Input, with the main difference being that
Integrated Gradients computes the average gradient as the input varies along a linear path from a baseline $\Tilde{x}$ to $x$. The baseline is defined by the user and often chosen to be zero. Please refer to Figure \ref{figs:various-back-propagation} for the mathematical definition for both methods.

Pixel-space gradient visualizations such as the above-mentioned Guided Back-propagation and Deconvolution are high-resolution and highlight fine-grained details in the image, but are not class-discriminative, i.e., the attribution value plots for different classes may look similar. In contrast, localization approaches like Class Activation Mapping (CAM)~\citep{zhou2016learning} are highly class-discriminative (e.g., the `cat' explanation exclusively highlights the `cat' regions but not `dog' regions in an image containing both a cat and dog). This approach modifies image classification CNN architectures by replacing fully-connected layers with convolutional layers and global average pooling, thus achieving class-specific feature maps. A drawback of CAM is that it requires feature maps to directly precede softmax layers, so it is only applicable to particular kinds of CNN architectures.
To solve this shortcoming, Grad-CAM~\citep{selvaraju2017grad} was introduced as a generalization to CAM, which uses the gradient information flowing into the last convolutional layer of the CNN to understand the importance of each neuron for a decision of interest. Furthermore, it is combined with existing pixel-space gradient visualizations to create Guided Grad-CAM visualizations that are both high-resolution and class-discriminative.

Besides gradients, back-propagation of activation values can also be leveraged as an interpretability approach. Layer-wise Relevance Propagation (LRP)~\citep{bach2015pixel} is the first to adopt this method, where the algorithm starts at the output layer $L$ and assigns the relevance of the target neuron equal to the output of the neuron itself (i.e., the activation value of the neuron) and the relevance of all other neurons to zero, as shown in Eq.~\ref{eq:LRP-eq-1}. Then the recursive back-propagation rule (called the $\epsilon$-rule) for the redistribution of a layer's relevance to the preceding layer is described in Eq.~\ref{eq:LRP-eq-2}, where we define $z_{ji}=w_{ji}^{(l+1,l)}x_i^{(l)}$ to be the weighted activation of a neuron $i$ onto neuron $j$ in the next layer and $b_j$ the additive bias of unit $j$. Once the back-propagation reaches the input layer, the final attributions are defined as $R_i^c(x)=r_i^{(1)}$. 
As an alternative, DeepLIFT~\citep{pmlr-v70-shrikumar17a} proceeds in a backward fashion similar to LRP but calibrates all relevance scores by subtracting reference values that are determined by running a forward pass through the network using the baseline $\bar{x}$ as input and recording the activation of each unit. 
Although LRP and DeepLIFT were invented based on back-propagation of activation values, it has been demonstrated in \cite{ancona2018towards} that they can also be computed by applying the chain rule for gradients and the converted equations are summarized in Figure \ref{figs:various-back-propagation}.

\begin{equation}
    r_i^{(L)}=
    \begin{cases}
      S_i(x) & \text{if unit}\ i\ \text{is the target unit of interest} \\
      0 & \text{otherwise}
    \end{cases}
    \label{eq:LRP-eq-1}
\end{equation}

\begin{equation}
    r_i^{l}=\sum_{j} \frac{z_{ji}}{\sum_{i'}({z_{ji'}+b_j})+\epsilon\cdot \mathrm{sign}(\sum_{i'}(z_{ji'}+b_j))}r_j^{l+1}
    \label{eq:LRP-eq-2}
\end{equation}

\subsection{Feature perturbation}
Compared to back-propagation based methods, which compute gradients of outputs with respect to input features, feature perturbation methods explicitly examine the change in model confidence resulting from occluding or ablating certain features. 

The idea of masking parts of the input and measuring the model's change in confidence was introduced in a model agnostic context, pre-DL by works such as \cite{trumbelj2009ExplainingIC} and \cite{robnik2008explaining}. Based on some of these works, there have been multiple methods in DL for feature perturbation, attempting to explain the model based on the change in output classification confidence upon perturbation of features. These include model agnostic works based on conditional multivariate analysis and deep visualization \citep{zintgraf2017visualizing} (based on the instance-specific method known as prediction difference analysis) and explicit erasure of parts of input representations \citep{li2016understanding}; as well as convolution neural network specific identification of image regions for which the model reacts most to perturbation \citep{matthew2014visualizing} and image masking models that are trained to manipulate scores outputted by the predictive model by masking salient parts of the input image \citep{dabkowski2017real}. Similar to image masking models, recent model-agnostic methods use generative models to sample plausible in-fills (as opposed to full masking) and optimize to find image regions that most change the classifier decision after in-filling \citep{chang2018explaining}. 

In the direction of more theoretically grounded variable importance-based techniques, \cite{fisher2019all} measure the model prediction difference upon adding noise to the features. Additionally, various adversarial perturbation techniques have been introduced that add noise to the feature representations, falling in the category of \textit{Evasion Attacks} \citep{tabassi2019taxonomy}. Evasion Attacks involve finding small input perturbations that cause large changes in the loss function and lead to mispredictions. These input perturbations are usually found by solving constrained optimization problems. These include gradient-based search algorithms like Limited-memory Broyden-Fletcher-Goldfarb-Shanno (L-BFGS) \citep{szegedy2013intriguing}, Fast Gradient Sign Method (FGSM) \citep{goodfellow2014explaining}, Jacobian-based Saliency Map Attack (JSMA) \citep{papernot2016limitations} and Projected Gradient Descent (PGD) \citep{madry2018attack} among others. For detailed surveys on adversarial perturbation techniques in computer vision see \cite{akhtar2018threat}; for surveys on adversarial attacks in general see \cite{chakraborty2018adversarial} and \cite{yuan2019adversarial}. While the goal of these methods is to actively change model confidence for the purpose of attacking the model, they take advantage of the black box nature of DL models and have led to creation of techniques that can be used to deploy more robust and interpretable models. 

\subsection{Attention}
Attention mechanisms have played an important role in model interpretations and the attention weights have been widely adopted as a proxy to explain a given model's decision making \citep{xu2015show, xie-etal-2017-interpretable,clark2019does,voita2019analyzing}.

Historically, attention mechanisms have been introduced in the context of sequence to sequence text model alignment as the way to directly incorporate the importance of the context to any given word representation. Each input word in a given context is represented by a weighted sum of the representations of other words. Naturally, the dynamic weights for each word can be interpreted as the contribution (or importance) of the words to a given word representation.

While the exact architecture of the attention-utilizing models differs from model to model, all of them make use of the set of computations known as the attention mechanism.
The basic building block of attention is a generalized trainable function \citep{bahdanau2014neural,vaswani2017attention}:
\begin{equation}
Attention (V,Q,K) = Score(W_{q}Q, W_{k}K) \odot W_{v}V 
\end{equation}
where $ Q$ and $K$ represent the context of a given element, $V$ the unmodified element contribution to the representation without the context being taken into an account, and the set of weights $W_k, W_q, W_v$  are the adaptable weights that represent the learned elements' contributions.

The output of the Score function is known as the attention weights and represents the contributions of the other elements of the input to the representation of the given element or sequence as a whole; in the naive interpretation setting, a high post-training attention weight of an input feature or a set of features corresponds to a higher importance of the given feature value in producing a prediction.


Note that this methods of producing interpretation is intrinsically linked to the model itself and constitutes a direct interpretation of the outputs of the parts of a given attention-utilizing model (attention scores) as an explanation for the prediction. Since attention scores are often computed over already pooled representations of the elements and sequences, the element scores do not necessarily represent the direct feature contributions \citep{jain2019attention, brunner2019identifiability,zhong2019fine}.

The majority of work on attention based interpretability has been in the general time-series processing field due to both the success of the attention-using models and the natural idea of viewing the contribution of the other elements of the sequence to the current state \citep{sezer2020financial, fawaz2019deep,wang2019review,ardabili2019deep}.

Fully-attention based models and attention based interpretations are also popular in natural language processing (NLP) due to the compositional nature of syntax and meaning \citep{wolf2020transformers}.

\subsection{Model distillation} \label{distillation}
Model distillation  (also known as Network distillation) is a model compression technique where a simpler model (student) is “taught” by a more complex model (teacher). While the original use of the technique focused on the improved performance or compactness of the student model \citep{hinton2015distilling}, it is important to note that if the simpler model is naturally interpretable, the transfer results in an “interpretable” model explaining the behavior of the more complex teacher model.  

A complex model's behavior may be approximated either locally, by fitting a simpler model around a given example to produce an explanation for a given point \citep{ribeiro2016should}, or globally, by fitting one simple model directly to the teacher model, using all the training data~\citep{lakkaraju2017interpretable}.

Due to the explicit “interpretation” use case of the technique, the student models are in general limited to either generalized linear models \citep{ribeiro2016should}, decision trees \citep{craven1995extracting,schmitz1999ann,plumb2018model} or direct rules or set  inductions \citep{sethi2012kdruleex,lakkaraju2017interpretable,ribeiro2018anchors,zilke2016deepred}

The most influential member of this family of interpretation producing techniques is LIME \citep{ribeiro2016should}, a general method for generating local explanation for a specific input case. The local model that serves as an explanation for a given point is obtained by minimizing 
\begin{equation}
\xi(x^*)= \argmin_{g \in \mathcal{G}} \mathcal{L}(f,g, \pi_{x^*}) + \Omega(g) 
\end{equation}
where $\mathcal{G}$ is a class of the interpretable models used to produce an explanation, $\pi_{x^*}$ defines the neighborhood of points near $x^*$, $\mathcal{L}$ is the measure of the difference between the original model and the explanation model prediction in that neighborhood, and $\Omega(g)$ is a complexity measure of the explanation model.

In practice, in the classical LIME use, $\mathcal{L}$ is set to be the distance weighted squared loss between the original model and the explanation model prediction computed over a randomly sampled set of data points biased to lie near $x^*$ by $ \pi_{x^*}$. The explanation model class $\mathcal{G}$ is the class of all linear models and $\Omega(g)$ is a regularization term to prevent overfitting. 

The vast majority of local knowledge distillation for interpretability models are the result of modifying Lime in either the neighborhood construction (ALIME \citep{shankaranarayana2019alime}), sampling (MPS-LIME \citep{shi2020modified}) and input structure constraining procedure (GraphLime \citep{huang2020graphlime})  or the nature of the explanation model (SurvLIME \citep{kovalev2020survlime}, GRAPHLime \citep{huang2020graphlime}). Another popular trend is producing semi-global explanation models through LIME-like fitting procedures (LIME-SUP \citep{Hu2018LocallyIM}, Klime \citep{hall2017machine}, NormLime \citep{ahern2019normlime}, DLIME \citep{zafar2019dlime}, ILIME \citep{elshawi2019ilime}).


\subsection{Game theory based Interpretability Methods} \label{gametheory}
DL models can also be interpreted via Shapley value, a game theory concept inspired by local surrogate models~\citep{lundberg2017unified}.
Shapley value is a concept of fair distribution of gains and losses to several unequal players in a cooperative game~\citep{shapley1953value}. 
It is an average value of all marginal contributions to all possible interactions of features (i.e., players in the game) given a particular example.
Therefore, the Shapley value can explain how feature values contribute to the model prediction of the given example by comparing against the average prediction for the whole dataset. 
Nevertheless, the Shapley value approximation is not easy to compute when the learning model becomes complicated.

Recently, researchers proposed a unified framework, SHAP (SHapley Additive exPlanations) values, to approximate the classical Shapley values with conditional expectations for various kinds of machine learning models, which include linear models, tree models~\citep{lundberg2018consistent}, and even complicated deep neural networks~\citep{lundberg2017unified}.
SHAP has been widely used recently for DL interpretation, yet researchers also admit to concerns about this popular interpretability method.

First, the SHAP for neural networks (KernelSHAP) is based on an assumption of model linearity. 
To mitigate the problem, \cite{ancona2019explaining} propose a polynomial-time approximation algorithm of Shapley values, Deep Approximate Shapley Propagation (DASP), to learn a better Shapley value approximation in non-linear models, especially deeper neural networks.
DASP is a perturbation-based method using uncertainty propagation in the neural networks. It requires a polynomial number of network evaluations, which is faster than other sampling-based methods, without losing approximation performance.
Also,~\cite{sundararajan2019many} show that SHAP, or other methods using Shapley values with conditional expectations, can be sensitive to data sparsity and yield counterintuitive attributions that make an incorrect model interpretation.
They propose a technique, Baseline Shapley, to provide a good unique result.

\subsection{Example based Interpretability Methods}
Instead of explaining the model using the attributive contribution of input data points, example based methods interpret the model behavior using only the particular training data points that are representative or influential for the model prediction.

For DL models, there are several interpretation methods based on example-level information.
For example, the influence function~\citep{koh2017understanding}, example-level feature selection~\citep{chen2018learning}, contextual decomposition (CD)~\citep{murdoch2018beyond}, and the combination of both prototypes and criticism samples---data points that can’t be represented by prototypes~\citep{kim2016examples}.
Other popular methods for interpretation, such as LIME~\citep{ribeiro2016should} (Section \ref{distillation}) and SHAP~\citep{lundberg2017unified} (Section \ref{gametheory}), also provide example-level model interpretability. 

The influence function is an example of example-based interpretability~\citep{koh2017understanding}, which can be used in both computer vision~\citep{koh2017understanding}, and NLP~\citep{han2020explaining}.
The goal of the influence function is to measure the change in the loss function as we add a small perturbation, weight, or remove a influence instance, which is a representative, influential training point.
Under the smoothness assumptions, the influence function can be computed using the inverse of the Hessian matrix of the loss function or by using the Hessian-vector products to approximate the result.
The influence function can also be used to generate an adversarial attack.

Researchers developed DL-based instance-wise feature selection at the example-level for feature importance measurement~\citep{chen2018learning}.
Instance-wise feature selection (L2X, Learning to Explain) measures feature importance locally for each specific example and therefore indicates which features are the key for the model to make its prediction on that instance.
L2X is trained to maximize the mutual information between selected features and the response variable, where the conditional distribution of the response variable given the input is the model to be explained.
To solve an intractable issue of direct estimation of mutual information and discrete feature subset sampling, the authors apply a variational approximation for mutual information, then develop a continuous reparameterization of the sampling distribution.
The method has been applied to CNN and hierarchical long short-term memory (LSTM) on different datasets and yields a better explanation performance quantitatively and qualitatively.

CD is an interpretation method to analyze individual predictions by decomposing the output of LSTMs without any changes to the underlying model~\citep{murdoch2018beyond}. 
In NLP, it decomposes an LSTM into a sum of two contributions: those resulting solely from the given phrase and those involving other factors. 
CD captures the contributions of combinations of words or variables to the final prediction of an LSTM. 
In the study, researchers demonstrate that CD can explain both NLP and general LSTM applications. 
For example, they model for sentiment analysis by identifying words and phrases of differing sentiment within a given review and extracting positive and negative words from the model. 
The CD method can be further extended to a more general version, contextual decomposition explanation penalization (CDEP)~\citep{rieger2019interpretations}.
CDEP is a method that allows the insertion of domain knowledge into a model to ignore spurious correlations, correct errors and generalize to different types of dataset shifts.
It is general and can be applied to different neural network architectures.

For graph neural networks,~\cite{ying2019gnn} further propose a model-agnostic GnnExplainer to provide interpretability on graph-based tasks, such as node and graph classification.
By identifying the prediction-relevant edges, GnnExplainer can highlight local subgraph structures and small subsets of important features to the prediction.
The method can be used for single and multiple instance explanations in a graph.

To tackle the real-world data, which may not have a set of prototypical examples representing the data well, we can also utilize both the prototypical examples and criticism samples that don’t fit the model well~\citep{kim2016examples}.
The MMD-critic (maximum mean discrepancy-critic) method uses a Bayesian approach to select the prototype and criticism samples and to provide explanations that can facilitate human reasoning and understanding of the model.

\subsection{Generative based Interpretability Methods}
The basis of generative based methods for explaining a model's behavior uses information that does not occur explicitly in attributes of the input, but is derived from external knowledge sources, from a causal model, or from explainable probabilistic modeling.

For example, the state-of-the-art general domain neural question answering (QA) system attempts to provide human-understandable explanations for better commonsense reasoning, yet to interpret how the model utilizes common sense knowledge, a common-sense explanation generation framework is required~\citep{rajani2019explain}.
Researchers collect human narrative explanations for common sense reasoning and pretrain language models~\citep{rajani2019explain}, which can generate explanations and be used concurrently with the QA system (Commonsense Auto-Generated Explanations (CAGE) framework).
They further transfer knowledge (generated explanations) to out-of-domain tasks and demonstrate the capacity of pretrained language models for common sense reasoning.

Generative Explanation Framework (GEF) is another hybrid generative-discriminative method that explicitly captures the information inferred from raw texts, generates abstractive, fine-grained explanations (attributes), and simultaneously conducts classification tasks. 
It can interpret the predicted classification results and improve the overall performance at the same time~\citep{liu2019towards}.
More specifically, the authors introduce the explainable factor (EF) and the minimum risk training (MRT) approach that learn to generate more reasonable explanations.
They pretrain a classifier using explanations as inputs to classify texts, then adopt the classifier to jointly train a text encoder by computing EF, which is the semantic distance between generated explanations, gold standard explanations, and inputs, and then minimizing MRT loss that considers both the distance between predicted overall labels and ground truth labels, as well as the semantic distance represented in EF.
GEF is a model-agnostic method that can be used in different neural network architectures.

\cite{madumal2019explainable} introduced action influence models that utilize the structural causal model to generate the explanation of the behavior of model-free reinforcement learning agents through knowing the cause-effect relationships using counterfactual analysis.
The proposed model has been evaluated on deep reinforcement learning algorithms, such as Deep Q Network (DQN)~\citep{mnih2013playing}, Double DQN (DDQN)~\citep{van2016deep}, Proximal Policy Optimization (PPO)~\citep{schulman2017proximal}, and Advantage Actor Critic (A2C)~\citep{mnih2016asynchronous}.

\cite{wisdom2016interpretable} developed a model-based interpretation method, sequential iterative soft-thresholding algorithm (SISTA), to construct recurrent neural network (RNN) without black-box components like LSTMs, via the trained weights of the explicit probabilistic model. 

\section{Methods for Interpretability in Healthcare}

In the last section, we have summarized the methodology for each class of interpretation methods. Most of these methods were initially proposed for general domain applications. In order to deploy them to healthcare problems, some customization needs to be performed. Therefore in this section, we discuss how each class of interpretation methods can be adapted to healthcare systems. We also discuss what kinds of clinical/medical observations and findings we can make with the help of these interpretation methods. 

\subsection{Back-propagation}
Back-propagation based interpretability methods have been widely used to help visualize and analyze those DL models adopted for healthcare problems, which include computer vision, NLP~\citep{gehrmann2018comparing}, time series analysis, and static features-based predictive modeling. 
We would like to summarize these successful applications and categorize them based on the applied task types.

In computer vision tasks, many powerful DL models have achieved close to expert doctor performance~\citep{esteva2017dermatologist,ran2020deep} and thus it is very meaningful to study how these models can accomplish such great success~\citep{singh2020explainable}. \cite{xie2019interpretable} adopted CAM~\citep{zhou2016learning} to generate heatmaps separately for melanoma and nevus cells in skin cancer histology images so that the morphological difference between these two types of cells can be visualized: the melanoma cells are of irregular shape and the nevus cells are distinctly shaped and regularly distributed. \myRed{\cite{zhang2021anc} used Grad-CAM to provide an explainable heatmap for an attention network built for classifying chest CT images for COVID-19 diagnosis, while Grad-CAM was also used to explain a graph convolutional network for secondary pulmonary tuberculosis diagnosis based on chest CT images ~\citep{wang2021explainable}.} Integrated Gradients~\citep{sundararajan2017axiomatic} was used to visualize the features of a CNN model used for classifying estrogen receptor status from breast magnetic resonance imaging (MRI) images~\citep{pereira2018automatic}, where the model was found to have learned relevant features in both spatial and dynamic domains with different contributions from both. Overall, back-propagation based methods have been used to visualize and interpret various medical imaging modalities such as brain MRI~\citep{eitel2019testing}, retinal imaging~\citep{sayres2019using,singh2020interpretation}, breast imaging~\citep{papanastasopoulos2020explainable,kim2018visually}, skin imaging~\citep{young2019deep}, computed tomography (CT) scans~\citep{couteaux2019towards}, and chest X-rays~\citep{linda2020tailored}. 

For features-based predictive modeling, back-propagation based interpretability methods can be applied to both static and time-series analysis. For static analysis (e.g., therapy recommendation based on a fixed set of features), fully connected neural networks are typically utilized for modeling and thus are the target to be interpreted. Commonly-used interpretability methods include DeepLIFT~\citep{fiosina2020explainable}, LRP~\citep{li2018leveraging,zihni2020opening}, etc. For time-series analysis, besides being able to analyze which features are more important or relevant to the prediction among all features used~\citep{yang2018explaining}, it is noteworthy that we can also analyze what temporal patterns are more influential to the final model decision~\citep{mayampurath2019combining,suresh2017clinical}. 

\subsection{Feature perturbation}
Feature perturbation methods have primarily been discussed in the context of adversarial attacks in the healthcare domain \citep{finlayson2019adversarial}, mainly as potential future risks due to the ready acceptance of machine learning in diagnosis and insurance claims approval. Nevertheless, the features that are most influential if altered by an attacker are also the ones to which the model's responses are most sensitive \cite{finlayson2019adversarial}. 


\cite{finlayson2018adversarial} perform adversarial perturbations (a variation on FGSM attack \citep{goodfellow2014explaining}) by addition of gradient-based noise to three highly accurate deep learning systems for medical imaging. By attacking models that classify diabetic retinopathy, pneumothorax and melanoma, they show vulnerabilities in three of the most highly visible successes for medical deep learning. In addition, they discuss hypothetical scenarios of how attackers could take advantage of the vulnerabilities the systems  demonstrate. More broadly, they comment on industries and scenarios that could be affected by adversarial attacks in the future: insurance fraud and determining pharmaceutical and device approvals. They discuss the challenging tradeoff between forestalling approval until a resilient algorithm is built and the harm that delaying the deployment of a technology impacting healthcare delivery for millions could entail. 

\cite{newaz2020attacks} show vulnerability in smart healthcare systems (SHS) by manipulating device readings to alter patient status. By performing two types of attacks, including Evasion Attacks \citep{tabassi2019taxonomy}, they identify flaws in an underlying ML model in a SHS. Employing feature perturbation methods such as FGSM \citep{goodfellow2014explaining}, randomized gradient-free attacks based on \cite{carlini2017towards},  \cite{croce2019scaling}, and  \cite{croce2018randomized} and zeroth order optimization based attacks \citep{chen2017zoo}, they are able to alter patient status for ML models based on patient vital signs. 

\cite{chen2020ecgadv} generate adversarial examples based on perturbation techniques for electrocardiograms. They use techniques from \cite{carlini2017towards} and \cite{athalye2018synthesizing} to misguide arrhythmia classification. In a similar application, \cite{han2020deep} introduce a smooth method for perturbing input features to misclassify arrhythmia. 

For temporal data in EHR, \cite{sun2018identify} introduce an optimization based attack strategy, similar to \cite{chen2017zoo}, to perturb EHR input data. \cite{an2019longitudinal} introduce a JSMA and attention-based attack by jointly modeling the saliency map and attention mechanism. 
Finally, in a domain agnostic setting, \cite{naseer2019cross} introduce cross-domain transferability of adversarial perturbations using a generative adversarial network (GAN)-based framework. They show how networks trained on medical imaging datasets can be used to fool ImageNet based classifiers. Successful transferability of adversarial perturbations can make it even simpler to fool healthcare models across multiple task domains, and potentially modalities. One such paper examines the effect of universal adversarial perturbations in the medical imaging space \citep{hirano2021universal}.  

\myRed{Several methods have been proposed in the general domain to counter these adversarial attacks, namely proactive defense (\cite{cisse2017parseval, gu2014towards, papernot2016distillation}, \cite{shaham2018understanding}) and reactive defense (\cite{feinman2017detecting, grosse2017statistical, lu2017safetynet}). Proactive defense methods increase the robustness of models retroactively, whereas reactive defense models detect the adversarial examples. There are also other methods, such as using collaborative multi-task learning (\cite{wang2020defending}). While it may seem that the possibility of adversarial perturbations works against the recommendation of using deep learning in healthcare settings, there are recent works pushing the boundaries by actively examining the reasons for the susceptibility of healthcare data to attacks (\cite{ma2021understanding}). Since feature perturbation techniques have strong policy-level implications in healthcare, it is also imperative to tailor general domain defense methods to the healthcare setting.}   




\subsection{Attention}
Attention architectures designed with a special consideration for interpretability are routinely used for EHR-based longitudinal prediction tasks such as heart failure prediction \citep{choi2016retain,kaji2019attention}, sepsis \citep{kaji2019attention}, intensive care unit (ICU) mortality \citep{shi2019deep}, automated diagnosis, and disease progression modeling \citep{gao2019camp, mullenbach2018explainable, ma2017dipole,bai2018interpretable,alaa2019attentive}. The underlying representation of such a model is often produced by an LSTM variant with the attention used to compute the contribution of a given feature or time step element of the sequence to the prediction. The best known model of this kind is RETAIN \citep{choi2016retain}, which includes computing the attention weights over both the time-step of the time-series and the individual features of the inputs. 


\myRed{Pure attention-based architectures such as the Transformer have revolutionized NLP-based modeling, allowing the use of massive unlabeled medical text for pretraining  \citep{lee2020biobert, alsentzer2019publicly,beltagy2019scibert}. Adoption of such models for non-text data is still relatively rare \citep{li2020behrt, rajan2017attend}.}

A special variant of the attention mechanism that seeks to address interpretability allows the model to output uncertainty on each input feature and use the aggregated uncertainty information for prediction~\citep{heo2018uncertainty}.

Despite the widespread use of the attention maps as explanations, we would caution against the direct interpretation of attention as an element's contributions to the prediction in the medical domain. More studies are needed to disentangle self-attention produced representations from the context contribution itself.

\subsection{Model distillation}
LIME \citep{ribeiro2016should} is one of the most popular techniques used to produce instance-level explanations for black-box model predictions in medical AI. The model-agnostic nature of the technique has led to its use in a diverse set of longitudinal EHR-based prediction tasks such as heart failure prediction \citep{khedkar2020deep}, cancer type and severity inferences \citep{moreira2020investigation}, breast cancer survival prediction \citep{hendriks2020machine}, and predicting development of hypertension  \citep{elshawi2019interpretability}.


It should be noted that the LIME variants are not widely used, despite the potential clinical usefulness of such interpretation methods.
Among the potentially useful variants for ML in medicine are SurvLIME \citep{kovalev2020survlime}, introduced specifically for  producing Cox proportional hazards explanations for black-box survival models, and DLIME \citep{zafar2019dlime}, a hierarchical clustering neighborhood based semi-global LIME variant for producing more consistent explanations for predictions over similar inputs.

\subsection{Game theory based Interpretability}
The game theory based SHAP algorithm has been widely applied in the medical domain for feature contribution analysis due to its ability to explain not only individual predictions but also global model behavior via the aggregation of Shapley values. 
SHAP is also model-agnostic, so that it can be applied to various machine learning algorithms~\citep{lundberg2017unified,lundberg2018explainable}.

In the direct usage of SHAP for deep learning in healthcare,~\cite{arcadu2019deep} applied SHAP to find the crucial regions, which are peripheral fields, for identifying diabetic retinopathy progression.
Also, for the interpretation of medical imaging, ~\cite{young2019deep} and \cite{pianpanit2019neural} utilized KernelSHAP to generate the saliency maps for interpreting the deep neural networks for melanoma prediction and Parkinson’s disease prediction, respectively.
\cite{levy2019pathflowai} also adopted SHAP to interpret the portal region prediction in pathology slide imaging.
Beyond medical imaging,~\cite{boshra2019neurophysiological} used SHAP to investigate the features’ influence on concussion identification given the electroencephalography (EEG) signals.

\cite{ancona2019explaining} uses the DASP algorithm to approximate the Shapley values and yields the explanation of the deep learning models and applies it to a fully-connected network model for predicting the Parkinson's disease rating scale (UPDRS), which is a regression task to predict the severity of Parkinson’s disease based on 18 clinical features in a telemonitoring dataset.

\myRed{In~\cite{lundberg2018explainable}, they also have anesthesiologists consulted to ensure that their model explanations are clinically meaningful. 
The anesthesiologists were asked to justify the SHAP explanations with the change in model output when a feature is perturbed. 
~\cite{li2020machine} also shows that it is possible to use SHAP for modeling and visualizing nonlinear relationship between prostate-specific antigen and Gleason score in prostate cancer that is consistent with the prior knowledge in the medical literature. 
Such clinical evaluations help the medical community to accept the interpretation method better.}

Other works mentioned in this section also provide explanations that are aligned with prior knowledge and ground truth given by the dataset via visualization or computing quantitative metrics, yet none of them are justified by a formal clinical user study. 
Further study is needed for these methods and applications in healthcare.

\myRed{One major concern of using SHAP in the medical domain is that the Shapley value and SHAP was originally derived from economics tasks, where the cost is additive.
However, clinical features are usually heterogeneous, and the Shapley values derived from the model may not be meaningful in the domain~\citep{kovalerchuk2021survey}.
Further investigation is needed to justify real-world clinical use of SHAP-based interpretations.}

\subsection{Example based Interpretability}
Example-based model interpretation provides a mental model that allows clinicians to refer to some similar cases, prototypes, or clusters given a new case.

Researchers utilize CDEP to ignore spurious confounders in skin cancer diagnosis~\citep{rieger2019interpretations}.
The study uses a publicly available image dataset from ISIC (International Skin Imaging Collaboration), which has colorful patches present in approximately 50\% of the non-cancerous images but not in the cancerous images. 
It can be problematic if the learned model uses such spurious patch features as an indicator but not the critical underlying information for skin cancer prediction. 
The CDEP helps penalize the patches for having zero importance during training and mitigates the issue.

\myRed{Although yielding better model performance with a quasi-explanation with a skin cancer classification example, CDEP has not yet been justified by a formal clinical user study and not yet been accepted by the medical community. 
It is still at the research rather than the deployment stage.}

\subsection{Generative based Interpretability}
DL interpretability can also be learned based on expert-interpretable features provided during the learning process.

To provide visually interpretable evidence for breast cancer diagnostic decisions, \cite{kim2018visually} developed an interpretability framework that includes a breast imaging reporting and data system (BIRADS) guided diagnosis network and a BIRADS critic network.
The interpretable 2D BIRADS guide map, which is generated from the visual feature encoder, can help the diagnosis network focus on the critical areas related to the human-understandable BIRADS lexicon via the critic network. 

\myRed{The study shows that with the BIRADS guide map, the performance is significantly higher than the network without the guide map. 
This finding also indicates the critical role and necessity of integrating medical domain knowledge while deploying machine learning models in healthcare.}

For radiology,~\cite{shen2019interpretable} proposed an interpretable deep hierarchical semantic convolutional neural network (HSCNN) for pulmonary nodule malignancy prediction on CT images.
HSCNN generates the binarized low-level expert-interpretable diagnostic semantic features that are commonly used by radiologists, such as sphericity, margin, and calcification; these are inputs to the high-level classification model, along with the latent representations learned from the visual encoder. 

\myRed{Both~\cite{kim2018visually} and~\cite{shen2019interpretable} demonstrate that the image guide map and label generation process may help clinicians curate the raw image information to high-level diagnostic criteria, yet the method is not yet justified by formal clinical user studies. 
Further study is needed for these methods to be accepted by the medical community.}

\section{Discussion}

\subsection{Dimensions of different interpretability methods} 
The current literature presents several different classification schemes for interpretation methods in DL \citep{lipton2018mythos, doshi2017towards, pedreschi2019meaningful}.
In addition to the methodology motivated classification used in the Interpretability methods section of the paper, we present two different questions that every interpretation producing method naturally poses:

\begin{enumerate} 
\item  \textbf{Model Dependence:} Does the explanation model depend on the internal structure of the model it is explaining or can it be used for producing an explanation of any ``black-box" model?
\item \textbf{Explanation Scope:} Does the explanation model focus on producing an explanation for a given input-prediction pair or is it attempting to create a unified global explanation of the model's behavior?
\end{enumerate}

A characterization of the most commonly used methods to produce interpretations in health care with respect to these aspects is presented in Table~\ref{tab:shared_clinical_tasks}. 

In general, the vast majority of the methods are explicitly local, producing the explanation for a given decision only, with some attempts at aggregation of the local explanation into patterns \citep{natesan2020model, lakkaraju2019faithful}.

The community appears to be deeply split on the issue of model dependence, with the proponents citing the necessity of explanation fidelity \citep{rudin2019stop}, while opponents doubt the inherent fidelity of the directly model-dependent explanations \citep{jacovi2020towards} and stress the need for flexible model-independent explanation methods \citep{ribeiro2016model}.



\begin{table}[p]
\small
    \centering
    \begin{tabular}
    {p{0.12\textwidth}p{0.15\textwidth}p{0.05\textwidth}p{0.05\textwidth}p{0.48\textwidth}p{0.05\textwidth}}
    \toprule
        Class:    & Model & Scope    & Dep.                      & Potential Issues & Ref. \\ \midrule
        \multirow{ 2}{*}{Back-prop.} & \href{https://github.com/ankurtaly/Integrated-Gradients}{Integrated Gradients} & L & I & More computationally expensive than Gradients * Inputs, the baseline needs to be carefully selected/tuned for some cases & \cite{sundararajan2017axiomatic}  \\
        & \href{https://github.com/ramprs/grad-cam}{CAM} & L & I & Label/class discriminative features revealed by this method may not be convincing and accurate for some data samples & \cite{zhou2016learning}  \\
        & \href{https://github.com/alewarne/Layerwise-Relevance-Propagation-for-LSTMs}{LRP} & L & I & \myRed{Initially proposed for interpreting multi-layer perceptions and hard to generalize well to more complex neural networks such as LSTM and Transformers} & \cite{bach2015pixel}  \\
\hline

 \multirow{ 2}{*}{Feat perturbation} & \href{https://github.com/lmzintgraf/DeepVis-PredDiff}{Prediction Difference Analysis} & L & I & Computationally expensive. Simulates the absence of feature via marginalization, rather than exact knowledge of model behavior without the feature present. & \cite{zintgraf2017visualizing}  \\
& \href{https://arxiv.org/pdf/1612.08220.pdf}{Representation Erasure} & G & I & Computationally expensive, requires several steps of probing. Injects random noise into input for representation erasure.  & \cite{li2016understanding}  \\ 
& \href{https://github.com/zzzace2000/FIDO-saliency}{Counterfactual Generation} & L & I & Computationally expensive due to intermediate generative stage for injecting noise respecting data distribution. Involves similar marginalization approximations as Prediction Difference Analysis. & \cite{chang2018explaining} \\ \hline

            \multirow{ 2}{*}{Attention} &  \href{https://github.com/mp2893/retain}
{RETAIN} & L & D & Attention weight correspond to the importance of the intermediate representations to the final representation, not the input elements directly.   & \citep{choi2016retain}    \\
 & \href{https://github.com/khirotaka/SAnD}{Attend and Diagnose} & L & D & Same issues as retain, exacerbated by the use of the fully-additional architecture of the model  & \citep{rajan2017attend}  \\ \hline

 \multirow{ 2}{*}{Model distillation} & \href{https://github.com/marcotcr/lime}{LIME} & L & D & The explanation is not cross instances consistent and might vary wildly for even for very similar instances. Possible issues on discontinuous data. Might produce inconsistent results across the multiple runs  & \cite{ribeiro2016should}  \\
& \href{https://github.com/marcotcr/anchor} {Anchors} & S & I & Might produce inconsistent results across multiple runs. The explanation might be too specific and not very robust at the decision boundary  & \citep{ribeiro2018anchors} \\ \hline
 \multirow{ 2}{*}{Game theory} & \href{https://github.com/slundberg/shap}{SHAP} & L & I & Computationally expensive. Require the access to training data for interpretation. & \citep{lundberg2017unified} \\
 \hline
 \multirow{ 2}{*}{Example} & \href{https://github.com/kohpangwei/influence-release}{Influence function} & L & I & Won't work for models without differentiable parameters and losses. Only an approximation. No clear cut of ``influential'' and ``non-influential''. May not be human-interpretable if there are too many feature values in a prototype. & \citep{koh2017understanding}  \\
& \href{https://github.com/jamie-murdoch/ContextualDecomposition}{Contextual decomposition} & L & D & Only for LSTM. Require further algorithm modifications to extend the method for other network architecture. & \citep{murdoch2018beyond} \\ \hline
 \multirow{ 2}{*}{Generative} & CAGE & L & D & Require high-quality external knowledge resources. Task-specific method. & \citep{rajani2019explain} \\ 
    \bottomrule \end{tabular}
    \caption{Most popular methods \myRed{intended for providing interpretation} (cited more than 50 times), Scope : \textbf{L}ocal, \textbf{S}emi-Global, \textbf{G}lobal.
    Model Dependence: \textbf{D}ependent, \textbf{I}ndependent}
    \label{tab:shared_clinical_tasks}
\end{table}

\subsection{Credibility and Trustworthiness of Interpretability Methods}
In this section we will discuss two aspects of the methods used to produce interpretations of decision models used in health care:
\begin{enumerate}
    \item How faithful is the interpretation to the underlying decision making model?
    \item How understandable are the interpretations to human expert users?
\end{enumerate}

The two aspects are often at odds with each other: A complex model decision might require a rather complex explanation to cover all of the possible aspects of the model's behaviors on different inputs, which might not look easy to understand to humans. 

\subsubsection{Faithfulness of the interpretation}
We first discuss the direct correspondence between the produced interpretation and the model's decision making, known in the literature under the terms Fidelity~\citep{jacovi2020towards} or Faithfulness~\citep{rudin2019stop}. A perfectly faithful interpretation accurately represents the decision making of the model being explained. If the interpretation is constrained to agree with the model's behavior on \textbf{all possible inputs}, then no simpler explanation than the original model is possible. Even model dependent explanation producing methods may not be faithful to the original model because, as a simplified model, they may not include all parts of the original decision making process \citep{jain2019attention}.

When using an explanation producing model for black-box models trained on complex healthcare data, we recommend the user to consider the following issues to gain more insight into the explanation model's faithfulness.
\begin{enumerate}
    \item For explanations that, in themselves, are predictive models, look at the prediction agreement between the explanation model and the original: if the concordance is low, then the model is not faithful.
    \item While it is hard to estimate the fidelity of an explanation method, consider computing recently proposed fidelity measures over the set of the explanation methods you are planning to use \citep{yeh2019fidelity}.
    \item Consider running ``feature occlusion'' sanity checks, to check if changing those model elements according to the explanation change the original predictions \citep{hooker2018evaluating}.
    \item Due to the nature of some interpretation producing models, the same model might produce different explanations for the same pair of input-outputs over multiple runs.
\end{enumerate}

\subsubsection{Plausibility of the interpretation as defined by the expert user}
Traditionally, clinicians tend to embrace expert-curated models, such as the APACHE (Acute Physiology and Chronic Health Evaluation) score for evaluating the patient severity in the ICU~\citep{knaus1985apache}, due to the consistency between used model features and domain knowledge.
In contrast, machine learning approaches for healthcare problems aim to further improve performance by learning a much more complex representations from raw features while sacrificing model transparency.
Machine learning interpretability methods may provide human-understandable explanations, yet it is crucial that the explanations should be aligned with our knowledge to be trustable, especially for real-world deployment in the healthcare domain.
However, current deployments with interpretability methods mainly focus only on helping to debug the model  for engineers, but not the real-world use for end users~\citep{bhatt2020explainable}.
The appropriate interpretability methods should be selected and evaluated both to help model developers (data scientists and machine learning practitioners) understand how their models behave, and to assist clinicians to understand the rationale for model predictions for decision making.

For model developers, researchers evaluate their use of interpretability methods with different levels of model transparency (generalized additive models (GAMs) and SHAP), from both quantitative (machine-learned interpretability) and qualitative (visualization) perspectives using interviews and surveys~\citep{kaur2020interpreting}.
The results, however, show that developers usually over-trust the methods and this may lead to their misuse, especially over-relying on their ``thinking fast (system 1)''~\citep{kahneman2011thinking} since the good visualization may sway human thought, but may not fully explain the behavior of the system and may be incorrectly interpreted by developers. Moreover, visualization sometimes is not able to be fully understood and interpreted correctly by the model developers.
The authors point out that developers usually just focus on superficial values for model debugging instead of using explanations to dig deeper into data or model problems.
They also enumerate the common issues faced by developers, which include missing values, data change over time, data duplication, redundant features, ad-hoc categorization, and difficulties of debugging the methods based on explanations.
The developers are also shown to be biased toward model deployment even after recognizing suspicious aspects of the models.

From a clinical perspective, it is necessary and critical to have clinically relevant features that align with medical knowledge and clinical practice~\citep{caruana2015intelligible}, while under-performing models may still be acceptable as long as the errors are explainable.
In \cite{tonekaboni2019clinicians}, the authors survey clinicians in the ICU and emergency department to understand the clinicians’ need for explanation, which is mainly to justify their clinical decision-making to patients and colleagues.

Depending on the problem scope, different levels of interpretability may be considered by clinicians.
\cite{elshawi2019interpretability} conduct a case study of the hypertension risk prediction problem using the random forest algorithm and explore the important factors with different model-agnostic interpretability techniques at either global or local-level interpretation. 
They find that different interpretability methods in general provide insights from different perspectives to assist clinicians to have a better understanding of the model behavior depending on clinical applications. 
Global methods can generalize over the whole cohort while local methods show the explanation for specific instances. 
Thus, applications such as the hypertension risk prediction problem may focus on global risk factors derived from either global interpretability methods, mainly non-DL based techniques such as feature importance and partial dependence plot, or the aggregation of local explainers (e.g. SHAP, LIME)~\citep{elshawi2019interpretability}, while disease progression prediction requires integrated interpretations at local, cohort-specific and global levels~\citep{ahmad2018interpretable}.

However, different interpretability methods may yield a different subset of clinically relevant important features due to their ways to obtain feature importance. 
For instance, SHAP, coefficient of regression models, and permutation-based feature importance may provide completely different interpretations even if they are all at the global level.
\myRed{With some clinical examples, researchers found that the local interpretation methods (LIME and SHAP) of the correctly predicted samples are in general intuitive and follow common patterns, yet for the incorrectly predicted cases (either false positive or false negative cases), these local methods can be less consistent and more difficult to interpret~\citep{elshawi2019interpretability}.
Nevertheless, the users may not be aware of the assumption of using the model and how it makes the decision: e.g., the additivity assumption of the SHAP algorithm. Interpretability can be quite subjective, and the computerized techniques for producing interpretations lack the interactivity that is often crucial when one human expert is trying to convince another~\citep{lahav2018interpretable}.}

Studies also show shortcomings of some interpretability methods while adopting them for real-world clinical settings~\citep{tonekaboni2019clinicians,elshawi2019interpretability}.
For example, the complex correlation between features in feature importance-based methods, the weak correlation between feature importance and learned attention weights for recurrent neural encoders~\citep{jain2019attention}, and the trade-off between performance and explainability for rule-based methods, are all potential problems of using global interpretability methods~\citep{tonekaboni2019clinicians}.
For local interpretability methods, researchers also show that clinicians can easily conclude the explanation at the feature-level using LIME, but the main problem is that the LIME explanation can be quite unstable, where patients with similar patterns may have very different interpretations~\citep{elshawi2019interpretability}. 
\myRed{Instead, the advantage of the Shapley value interpretation method is that it makes the instance prediction considering all feature values of the instance, and therefore the patients with similar feature values will also have similar interpretations~\citep{elshawi2019interpretability}.}
But the cons of Shapley value-based methods are that they can be computationally expensive and that they need to access the training data while building model explainers~\citep{lundberg2017unified,janzing2020feature}.

It is not trivial to select appropriate interpretability methods for real-world healthcare applications.
Researchers therefore provide a list of metrics, including identity, stability, separability, similarity, time, bias detection and trust, to evaluate different interpretability methods when considering real-world deployment~\citep{elshawi2020interpretability}. 
However, they find that there is no consistent winning method for all metrics across various interpretability methods, such as LIME, SHAP and Anchors. 
Thus, it is essential to make a clear plan and think more about the clinical application and interpretability focus in order to select the reasonable and effective interpretability methods and metrics for real-world use. 

To further achieve the potential clinical impact of deployed models, we should not only focus on advancing machine learning techniques, but also need to consider human-computer interaction (HCI), which investigates complex systems from the user viewpoint, and propose better designs to bridge the gap between users and machines.
End users' involvement in the design of machine learning tools is also critical to understand the skills and real needs of end users and how they will utilize the model outputs~\citep{ahmad2018interpretable,feng2019can}.
\cite{kaur2020interpreting} suggest that it may be beneficial to design interpretability tools that allow back-and-forth communication (human-in-the-loop) to make interpretability a bidirectional exploration, and also to build tools that can activate thinking via ``system 2'' for deeper reasoning~\citep{kahneman2011thinking}.

 \subsection{Benchmarking Interpretation Methods}
\myRed{Now we have many different kinds of interpretation methods to choose when we want to analyze a neural model, although they are still in need of further improvement.} At the current state of the art, which method we should choose still does not have a definite answer. The choice of the right interpretation method should depend on the specific model type we want to interpret; however, such a detailed and comprehensive guideline for all kinds of models to be analyzed is currently not available. Several recent studies started to look into this problem by benchmarking some popularly used interpretation methods applied to some neural models such as CNN, RNN, and transformer. For example, \cite{arras-etal-2019-evaluating} first use four interpretation methods, namely LRP, Gradient*Input, occlusion-based explanation~\citep{li2016understanding}, and CD~\citep{murdoch2018beyond}, to obtain the relevance scores of each word in the text for the LSTM model for text classification tasks, and then measure the change of accuracy after removing two or three words in decreasing order of their relevance. By comparing the percentage of accuracy decrement, they observe that LRP and CD perform on-par with the occlusion-based relevance, with near 100\% accuracy change, followed by Gradient*Input which leads to only 66\% accuracy change. This experiment indicates that LRP, CD, and occlusion-based methods can better identify the most relevant words than Gradient*Input. As a counterpart, \cite{ismail2020benchmarking} argue that one should not compare interpretation methods solely on the loss of accuracy after masking since the removal of two or three features may not be sufficient for the model to behave incorrectly. Instead, they choose to measure the precision and recall of features identified as salient by comparing against ground truth important features and report the weighted precision and recall as the benchmarking metric. However, their annotations of which features are important are synthesized rather than collected by human annotation, which is not that convincing. In a more theoretical way, \cite{bhatt2020evaluating} propose several equations as quantitative evaluation criteria to measure and compare the sensitivity, faithfulness, and complexity of feature-based explanation methods. 

Through these benchmarking evaluations, we find that different interpretation methods may vary a lot in their advantages and disadvantages. To make use of this fact, some studies propose to aggregate two kinds of interpretation methods so that they can complement each other~\citep{ismail2020benchmarking}. For instance, \cite{bhatt2020evaluating} develop an aggregation scheme for learning combinations of various explanation functions, and devise schemes to learn explanations with lower complexity and lower sensitivity. We hope to see more efforts along this direction to generalize such an aggregation scheme to a broader range of interpretation methods.

\section{Conclusion}
In this review, we provided a broad overview of interpretation methods for interpreting the black-box DL models deployed for healthcare problems. We started by summarizing the methodologies of seven classes of interpretation methods in Section 2. Then we proceeded to discuss how these methods, which were initially proposed for general domain applications, are adapted for solving healthcare problem in Section 3. Finally in Section 4, we continued discussing three important aspects in the process of applying these interpretation methods to medical/clinical problems: 1. Are these interpretation methods model agnostic? 2. How good are their credibility and trustworthiness? 3. How to compare the performance of the methods so as to choose the most appropriate one for use? We hope these summaries and discussions can throw some light onto the field of explainable DL in healthcare and help healthcare researchers and clinical practitioners build both high-performing and explainable models.

\section*{Funding Information}
The authors' work was supported in part by collaborative research agreements with IBM, Wistron, and Bayer Pharmaceuticals, and by NIH grant 1R01LM013337 from the National Library of Medicine. The authors declare no conflicts of interest.

\begin{table}[p]
\small
\centering
\begin{tabular}{ll}
\toprule
Abbreviation & Full Form                                          \\
\hline
A2C          & Advantage Actor Critic                             \\
AI           & Artificial Intelligence                            \\
BIRADS       & Breast Imaging Reporting And Data System           \\
CAGE         & Commonsense Auto-Generated Explanations            \\
CAM          & Class Activation Mapping                           \\
CD           & Contextual Decomposition                           \\
CDEP         & Contextual Decomposition Explanation Penalization  \\
CNN          & Convolutional Neural Network                       \\
CT           & Computed Tomography                                \\
DASP         & Deep Approximate Shapley Propagation               \\
DDQN         & Double Deep Q Network                              \\
DL           & Deep Learning                                      \\
DNN          & Deep Neural Network                                \\
DQN          & Deep Q Network                                     \\
EEG          & Electroencephalography                             \\
EF           & Explainable Factor                                 \\
EHR          & Electronic Health Record                           \\
EU           & European Union                                     \\
FGSM         & Fast Gradient Sign Method                          \\
GAM          & Generalized Additive Models                        \\
GAN          & Generative Adversarial Network                     \\
GDPR         & General Data Protection Regulation                 \\
GEF          & Generative Explanation Framework                   \\
HCI          & Human-Computer Interaction                         \\
HSCNN        & Hierarchical Semantic Convolutional Neural Network \\
ICU          & Intensive Care Unit                                \\
ISIC         & International Skin Imaging Collaboration           \\
JSMA         & Jacobian-based Saliency Map Attack                 \\
L-BFGS       & Limited-memory Broyden-Fletcher-Goldfarb-Shanno    \\
L2X          & Learning to Explain                                \\
LIME         & Local Interpretable Model-agnostic Explanations    \\
LRP          & Layer-wise Relevance Propagation                   \\
LSTM         & Long Short-Term Memory                             \\
ML           & Machine Learning                                   \\
MRI          & Magnetic Resonance Imaging                         \\
MRT          & Minimum Risk Training                              \\
NLP          & Natural Language Processing                        \\
PGD          & Projected Gradient Descent                         \\
PPO          & Proximal Policy Optimization                       \\
QA           & Question Answering                                 \\
ReLU         & Rectified Linear Unit                              \\
RETAIN       & Reverse Time Attention Model                       \\
RNN          & Recurrent Neural Network                           \\
SHAP         & Shapley Additive Explanations                      \\
SHS          & Smart Healthcare Systems                           \\
SISTA        & Sequential Iterative Soft-Thresholding Algorithm   \\
\bottomrule
\end{tabular}
\caption{Glossary of abbreviations and acronyms.}
\label{tab:glossary}
\end{table}

\bibliographystyle{apalike}
\bibliography{references}

\end{document}